\documentclass[conference]{IEEEtran}
\IEEEoverridecommandlockouts
\usepackage{cite}
\usepackage{amsmath,amssymb,amsfonts}
\usepackage{algorithmic}
\usepackage{graphicx}
\usepackage{textcomp}
\usepackage{xcolor}
\usepackage{booktabs}
\usepackage{hyperref}
\usepackage{stfloats}

\def\BibTeX{{\rm B\kern-.05em{\sc i\kern-.025em b}\kern-.08em
    T\kern-.1667em\lower.7ex\hbox{E}\kern-.125emX}}
\begin{document}

\title{\huge Enhancing Cross-Modal Contextual Congruence for Crowdfunding Success using 
Knowledge-infused Learning} 


\author{
    Trilok Padhi\(^{1}\), Ugur Kursuncu\(^{1}\), Yaman Kumar\(^{2}\), Valerie L. Shalin\(^{3}\), Lane Peterson Fronczek\(^{1,4}\) \\
    \textit{ \(^{1}\)Georgia State University, 
    \(^{2}\)Adobe MDSR, 
    \(^{3}\)Wright State University,
    \(^{4}\)California Polytechnic State University} \\
    tpadhi1@student.gsu.edu, ugur@gsu.edu, ykumar@adobe.com, valerie.shalin@wright.edu, lfroncze@calpoly.edu
}

\maketitle

\begin{abstract}
The digital landscape continually evolves with multimodality, enriching the online experience for users. Creators and marketers aim to weave subtle contextual cues from various modalities into congruent content to engage users with a harmonious message. This interplay of multimodal cues is often a crucial factor in attracting users' attention. However, this richness of multimodality presents a challenge to computational modeling, as the semantic contextual cues spanning across modalities need to be unified to capture the true holistic meaning of the multimodal content. This contextual meaning is critical in attracting user engagement as it conveys the intended message of the brand or the organization. In this work, we incorporate external commonsense knowledge from knowledge graphs to enhance the representation of multimodal data using compact Visual Language Models (VLMs) and predict the success of multi-modal crowdfunding campaigns. Our results show that external knowledge commonsense bridges the semantic gap between text and image modalities, and the enhanced knowledge-infused representations improve the predictive performance of models for campaign success upon the baselines without knowledge. Our findings highlight the significance of contextual congruence in online multimodal content for engaging and successful crowdfunding campaigns.
\end{abstract}


\begin{IEEEkeywords}
Multimodal Learning, Crowdfunding, Knowledge Graphs, Contextual congruence, User engagement
\end{IEEEkeywords}

\section{Introduction}
\label{sec:intro}
Contemporary online multimodal platforms provide an appealing, rich user experience from social media to crowdfunding and e-commerce. Users may promote and market their ideas, products, or personal brands through a combination of different modalities of data, such as text and images, to attract user attention. However, the properties of effective online multimodal marketing campaigns challenge computational analysis. Potentially, synergy across the multi-sensory information space facilitates faster and more accurate processing to establish semantic relationships \cite{laurienti2004semantic,spence2011crossmodal,wickens2008multiple}. While the computational modeling of multimodal content has shown advancements, there remains an opportunity for deeper integration of explicit human knowledge, experience, and reasoning. Earlier and smaller (multimodal) Visual Language Models (VLMs) such as MMBT \cite{kiela2020supervised}, ViLBERT \cite{lu2019vilbert}, and LXMERT \cite{tan2019lxmert} primarily focused on exploiting unimodal cues (e.g., textual, visual) to establish connections between image and text. More recent very large VLMs, such as LLaVA \cite{liu2023visual}, BLIP2 \cite{li2023blip}, and GPT-4 \cite{openai2023gpt4}, have demonstrated considerable progress in capturing \emph{implicit} multimodal semantic relationships. However, these models typically operate without semantic constraints, making them vulnerable to hallucinations--- \emph{superficially congruent} \cite{hasanyou,Meyer2022-cf} and  \emph{catastrophically incorrect} \cite{shuster2021retrieval,Maynez2020-cj}. Despite their advancement, these models often remain unable to identify the explicit semantic connections between modalities that are crucial for mirroring human interpretation. \cite{ji2020leveraging}. This limitation can pose challenges while computationally assessing and generating multimodal content for online marketing campaigns, highlighting the need for further enhancement of these models to harness their potential \cite{mcguffie2020radicalization,weidinger2022taxonomy,tamkin2021understanding,de2023chatgpt}. 

\begin{figure}[hbt!]
  \centering
  \includegraphics[width=8.5cm]{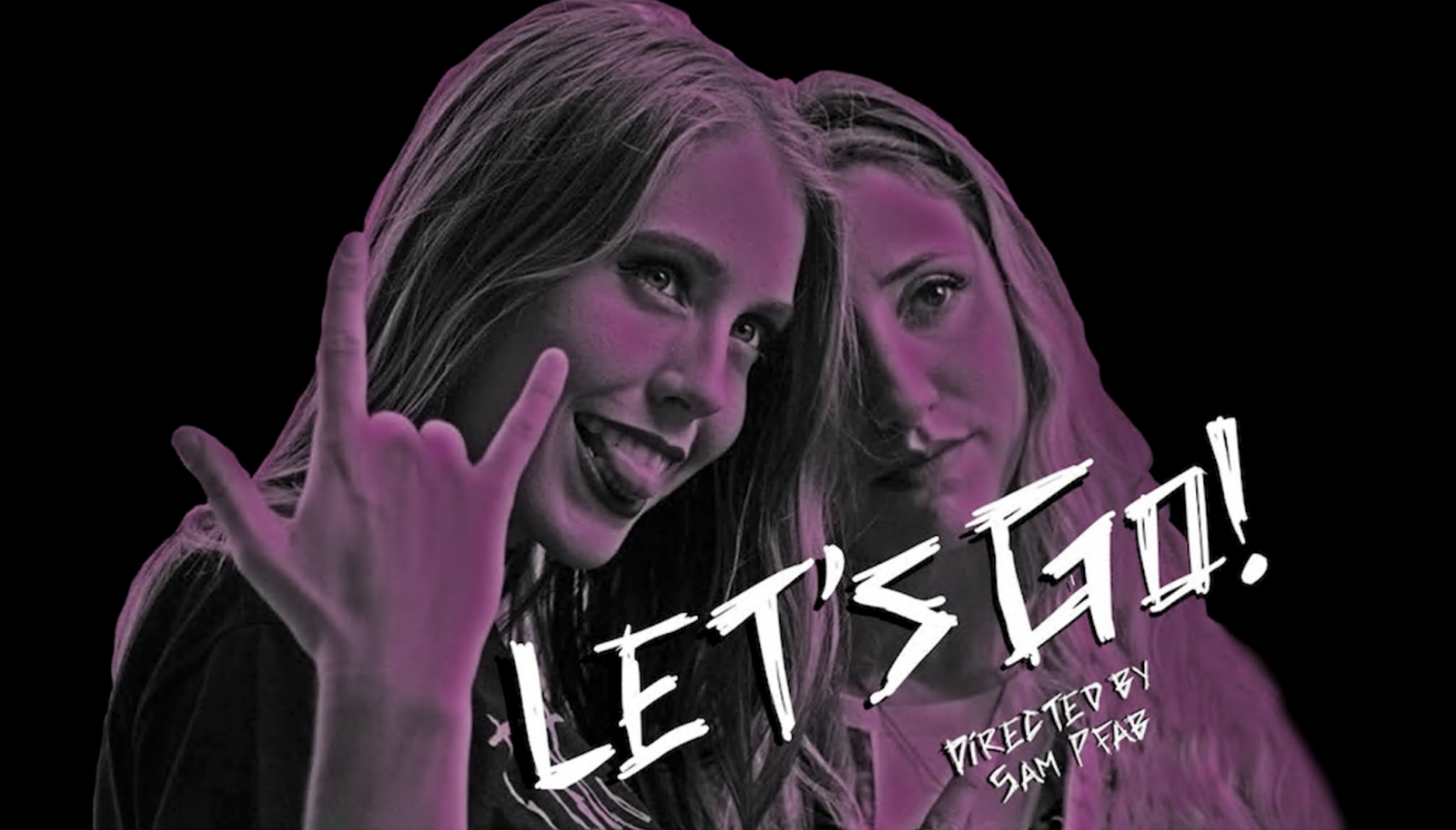}
\vspace{-0.5em}
\caption{An example pair of image and actual caption text from our crowdfunding dataset. The caption generated by the BLIP model and the actual human-generated caption of the image is below. 
\textbf{BLIP caption:} "Two women smiling with a hand gesture of rock and roll." 
\textbf{Actual human caption:} "Smash the glass ceiling. Destroy the patriarchy. Save the record store."}
\vspace{-0.5em}
\label{fig:image-womens-movement-patriachy}
\end{figure}

Consider the image and text in Figure \ref{fig:image-womens-movement-patriachy}, drawn from our dataset. 
The image contains two women smiling with the saga sign and the text \emph{"Let's Go!"}. The intended holistic meaning can be determined only after seeing the actual caption \emph{"Smash the glass ceiling. Destroy the patriarchy. Save the record store."} The creator of this marketing campaign seemingly intended to promote women's representation in society and the music industry by using an image of two assertive and confident women in the image. The pictured women were intended by the creator to represent the unstated subjects of smashing and, as such, challenging male dominance enshrined in patriarchy. Hence, to capture the correct meaning of such pairs of text and images, it is crucial to see both modalities and form contextually informed connections.

Table \ref{tab:generated-caption-two-women} presents explanations generated by state-of-the-art large VLMs, including BLIP2 and LLaVA. While LLaVA \cite{liu2023visual} did generate contextually informed explanations of the input pair, it produced hallucinations, such as describing, “The image features two women standing next to each other, \emph{both holding their cell phones in their hands. They seem to be taking selfies},” when, in fact, there are no cell phones visible in the given image, nor are they taking selfies. To address this issue, we appended the input pairs in our dataset with the most relevant concepts from ConceptNet {\cite{speer2018conceptnet}}. This augmentation mitigated hallucinations as demonstrated in the generated outputs from example pairs in Table \ref{tab:generated-caption-two-women}). This comparison highlights the value of multimodal analysis guided by external knowledge, which helps form more accurate cross-modal contextual connections. 
In this study, we examine the following two research questions (RQs): 

\textbf{RQ1:} Can we enhance the cross-modal contextual congruence of the representations of multimodal content by incorporating external knowledge by learning to unveil subtle cross-modal semantic relationships? 

\textbf{RQ2:} Do more contextually congruent representations help the models obtain more consistent and reliable predictive performance for the multimodal crowdfunding campaign success? 

\vspace{-1em}

Our approach enhances the cross-modal contextual congruence of representations from \emph{compact VLMs} for crowdfunding campaigns by incorporating an external Knowledge Graph (KG) (e.g., ConceptNet). As contextual congruence is defined as the association between multiple cues from various modalities (see Section \ref{sec:prelim}), we utilize ConceptNet, a commonsense KG with rich semantic knowledge of the world, to synthesize these diverse multimodal cues in a way that mirrors human cognitive processes. In this work, we first generated embeddings of image-text pairs from our dataset using text (e.g., language models) and image encoders (e.g., LVMs). Then, the captions for images were generated using BLIP \cite{li2022blip} to query the KG. We utilize semantic search to retrieve the most relevant concepts from ConceptNet and learn KG embeddings (KGEs) to represent these concepts. Initially, we learn multimodal data representations using bidirectional transformers. These representations are then fused with the KGEs through a fusion mechanism involving linear and multi-head cross-attention layers, resulting in enhanced knowledge-infused multimodal representations. By integrating external knowledge into the multimodal learning process, our approach aims to better capture the holistic meaning inherent in crowdfunding campaign content, ultimately improving the predictive performance of compact VLMs.

Our findings indicate that knowledge-infused multimodal representations consistently outperform baseline models without external knowledge. Further, we found that augmenting multimodal models with external knowledge bridges the semantic gap between modalities, enhancing their congruence by bringing them closer. Error analysis (see Section \ref{sec:results-discuss}) reveals that this incorporation of knowledge significantly improves the model's ability to capture the intended holistic, multimodal meaning. This study demonstrates the contribution of knowledge to machine learning algorithms for computational analysis of crowdfunding campaigns, which provide an ecologically valid measure of success. The contributions of this work lie in modeling a social and behavioral problem, such as successful crowdfunding campaigns, utilizing a neurosymbolic approach \cite{sap2022neural}. In so doing, we particularly tackle the problem of learning a contextually enhanced neural representation of multimodal content with external commonsense KG (i.e., symbolic) \cite{yu2022probabilistic}. This integration enables the model to better capture holistic meaning across modalities, thereby improving the performance in predicting the success of multimodal crowdfunding campaigns.

\makeatletter
\def\@makecaption#1#2{\vskip\abovecaptionskip \small #1. #2\par \vskip\belowcaptionskip}
\makeatletter

\begin{table*}[hbt!]
\vspace{-0.2em}
  \centerline{
    \begin{tabular}{p{0.8cm}p{1.3cm}p{14.6cm}}
    \toprule[2pt]
   \textbf{\centering Model} & \textbf{\centering Knowledge} & \textbf{Generated Caption} \\
    \toprule[1pt]
    BLIP2 & \centering No & the cover of let's go with two girls posing in front of a purple background going for a protest. \\
    \toprule[1pt]
    BLIP2 & \centering Yes & The cover of a movie "let's go" talks about patriarchy in the record store. \\ 
    \toprule[1pt]
    LLaVA & \centering No & The image features two women standing next to each other, both holding their cell phones in their hands. They seem to be taking selfies or sharing a moment together. One of the women is making a funny face, adding a playful vibe to the picture. The other woman is simply enjoying the moment and being part of the fun. \\ 
    \toprule[1pt]
    LLaVA & \centering Yes & The image features two beautiful and majestic black women standing together, united in their quest to smash the glass ceiling and destroy the patriarchy. They are surrounded by record albums, highlighting their shared passion for music and the record store. Their powerful presence and determination to save the record store from any threat make for a compelling scene that showcases their strength and resilience in the face of oppressive societal norms. \\
    \bottomrule[2pt]
    \end{tabular}%
  }
\caption{For the image and text pair presented in Figure \ref{fig:image-womens-movement-patriachy}, the captions were generated by large VLMs (BLIP2 and LLaVA), to describe the intended message. The image and text (actual caption) were used in prompts, with and without external knowledge. The knowledge was retrieved from ConceptNet as explained in section \ref{sec:explicit-knowledge} and appended to the prompts. The generated captions with knowledge show less hallucination based on our exploratory experiments with random samples from our dataset.}
\label{tab:generated-caption-two-women}
\vspace{-2em}
\end{table*}

\section{Background \& Related Work}
\label{sec:related}
\vspace{-1mm}
Recent multimodal VLMs, including LLaVA \cite{liu2023visual}, GPT-4 \cite{openai2023gpt4}, BLIP2 \cite{li2023blip}, BLIP \cite{li2022blip}, CLIP \cite{radford2021learning} have demonstrated significant success in capturing content, albeit with certain limitations, such as hallucinations \cite{li2023evaluating,cirik2022holm}. Earlier compact VLMs, including ViLBERT \cite{lu2019vilbert}, LXMERT \cite{tan2019lxmert}, and MMBT \cite{kiela2019supervised} introduced two-stream architectures that fused representations from pretrained unimodal models in later layers. Initial research explored text-image relations aiming to construct a holistic congruent meaning from word-object pairs \cite{wu2014multimodal,manning1998understanding}. 
As social media became prevalent, multimodal classifiers were developed to utilize text-image relations using LDA models \cite{wang2014bilateral,chen2013understanding}. More recent research expanded these classification systems to include image specificity \cite{jas2015image}, emotion \cite{chen2015velda}, interrelation metrics \cite{henning2017estimating,otto2020characterization}, and contextual and semiotic relations \cite{kruk2019integrating}, by using supervised learning to predict \emph{cross-modal correlations} \cite{chen2015velda,henning2017estimating,otto2020characterization,zhang2018equal,kruk2019integrating,vempala2019categorizing,kursuncu2018s}. Despite such progress, defining and classifying these correlations remain challenging due to the need for fine-grained annotations.

\vspace{-2mm}
\subsection{Knowledge-Infused Multimodal Learning}
\label{sec:knowledge-infusion-mm-learning}
\vspace{-1mm}
Large pretrained foundation models have become the dominant paradigm in vision and language modeling. Despite training on large datasets, such as Conceptual Captions \cite{Sharma2018-qk}, COCO \cite{Ren2015-oq}, Visual Genome \cite{Krishna2017-xz}, and ImageNet \cite{deng2009imagenet}, they suffer from certain limitations, including capturing surface-level characteristics and producing hallucinations. To address these challenges, researchers proposed to utilize KGs to enhance these models for various tasks, including Language Modeling \cite{yasunaga2022deep,zhang2022greaselm,Logan2019-tk,Lu2021-xe,kursuncu2020kil}, Natural Language Inference (NLI) \cite{Ni2021-db}, Language Generation \cite{Yu2022-uv,peters2019knowledge}, leading to developments of E-BERT \cite{Wu2022-ah}, ERICA \cite{Qin2020-et} ERNIE \cite{Zhang2021-ov}, ERNIE-NLI \cite{Bauer2021-rv}, and LUKE \cite{Yamada2020-zc}. To encode the complex relational data in multimodal content, composition-based multi-relational graph convolutional networks \cite{vashishth2020composition} were utilized.
Integrating external knowledge into VLMs has gained significant attention for visual commonsense reasoning (VCR). For instance, researchers have improved visual understanding and reasoning by combining scene graphs with structured knowledge from ConceptNet \cite{Zareian2020-vw}. However, many of the prior multimodal knowledge-enhanced approaches \cite{Garderes2020-gx,Marino2021-sn,Wu2022-lj} relied on keyword-based entity matching for explicit knowledge extraction, which may not capture deeper semantic relationships.
Knowledge-infused techniques have also been employed to enhance scene representations in autonomous driving \cite{wickramarachchi2022knowledge,wickramarachchi2021knowledge}. 
Additionally, knowledge-enhanced models have been effective in mitigating hallucinations in language and vision models, especially for tasks like Visual Question Answering (VQA) \cite{Chen2023-mf}. 
Surveys on the representation and application of KGs \cite{zou2020survey,ji2022survey} outline diverse applications of KGs, while recent work on reasoning over various graph types highlights the adaptability of KG-based reasoning for multimodal contexts \cite{chen2020survey}. 

\subsection{Modeling Crowdfunding Campaigns}
\label{sec:modeling-social-media}
\vspace{-1mm}
Crowdfunding platforms, such as Indiegogo and Kickstarter, seek venture capital from the general public. Between 2009 and 2023, over seven billion dollars have been pledged for the campaigns on Kickstarter. Some of the emerging brands such as Oculus \cite{Carradini2023-ka} originated from these platforms. These campaigns have a relatively low success rate, such as \%38.76 for the Kickstarter \cite{Carradini2023-ka}. While successful campaigns featured more images and links, the majority of prior crowdfunding studies focused primarily on textual content. Researchers \cite{Mitra2014-ly} analyzed the textual data in Kickstarter campaigns, extracting linguistic features with phrases (i.e., n-grams) and other control variables using a logistic regression model. Another study characterized the persuasive strategies in crowdfunding (Kiva) campaigns using semi-supervised textual RNNs \cite{yang2019let}. Researchers \cite{cheng2019success} developed one of the few predictive models for multimodal crowdfunding campaign success (MDL-TIM), which we use as a baseline in our study. 
The crowdfunding application provides an ideal domain for addressing the contextual congruence between modalities, providing both ample data and a measure of successful interpretation.

In contrast to this prior work, we use commonsense knowledge to reduce the semantic gap between modalities to enhance cross-modal congruence. Our work differentiates itself by jointly reasoning over both implicit (data-driven) and explicit (knowledge-driven) information retrieved through semantic search and infused into the model. This provides more contextually congruent representations and better performance in our downstream task. Further, to our knowledge, this is the first study applying KGs with language or visual models for crowdfunding, whereas prior approaches involving KGs used crowdsourcing \cite{kou2022crowdgraph,basharat2014semantically}.

\section{Preliminaries}
\label{sec:prelim}
\vspace{-1mm}
Multimodal marketing materials are often strategically designed to promote products, brands, individuals, or broadly entities, aiming to persuade audiences and encourage specific actions. In crowdfunding, the desired action is for users to commit as "backers," providing financial support to the campaign before its deadline. The decision-making process for potential backers involves synthesizing and evaluating both textual and visual content to form cross-modal semantic connections that capture the holistic message. 
These semantic connections facilitate a contextually congruent information environment that enhances cognitive processing, enabling humans to process information more effectively \cite{laurienti2004semantic,spence2011crossmodal}; thus, increasing the persuasiveness of the campaign message \cite{schubert2021multimodal}. 

\subsection{Cross-Modal Contextual Congruence}
\label{sec:context-congruence}
\vspace{-1mm}
Contextual congruence refers to associations between sensory cues across modalities, enhancing human response accuracy \cite{laurienti2004semantic,spence2011crossmodal}. Research indicates that cross-modal contextual congruence facilitates the selection and processing of semantically related information from different modalities \cite{almadori2021crossmodal,laurienti2004semantic,letts2022evaluating}, critical for predicting human response to multimodal crowdfunding campaigns. However, multimodality challenges computational modeling, as the semantic cues span across modalities to construct meaning. Our approach aims to enhance such cross-modal contextual congruence by infusing external knowledge into multimodal representations.

\section{Exploratory Analysis}
\label{sec:exploratory}
\vspace{-1mm}
We explore contextual congruence by examining the semantic distance between text and image and assessing whether external knowledge from ConceptNet reduces this distance. More specifically, we measured cross-modal contextual congruence by computing the semantic distance (i.e., cosine similarity) between image and text (actual caption) representations \cite{mandera2017explaining,frank2017word,beck2011congruence,maki2004semantic}. Our goal is to demonstrate how external knowledge can bridge semantic gaps between modalities, leading to potential improvements in predicting crowdfunding success. Captions for images were generated using BLIP, providing textual descriptions for each modality. Using semantic search (see Section \ref{sec:explicit-knowledge}), we retrieved most relevant concepts, which were appended to the text and the generated image captions. Then, we used BERT to create their representations. Clustering analysis with DB-SCAN was performed over pairs of text and generated image captions with and without knowledge, where we validated distinct clusters were formed for text and image modalities. Then, we analyzed how cluster centroids shifted with and without external knowledge to illustrate the impact of cross-modal contextual congruence.

\begin{figure}[hbt!]
  \centering
  \includegraphics[width=8.4cm]{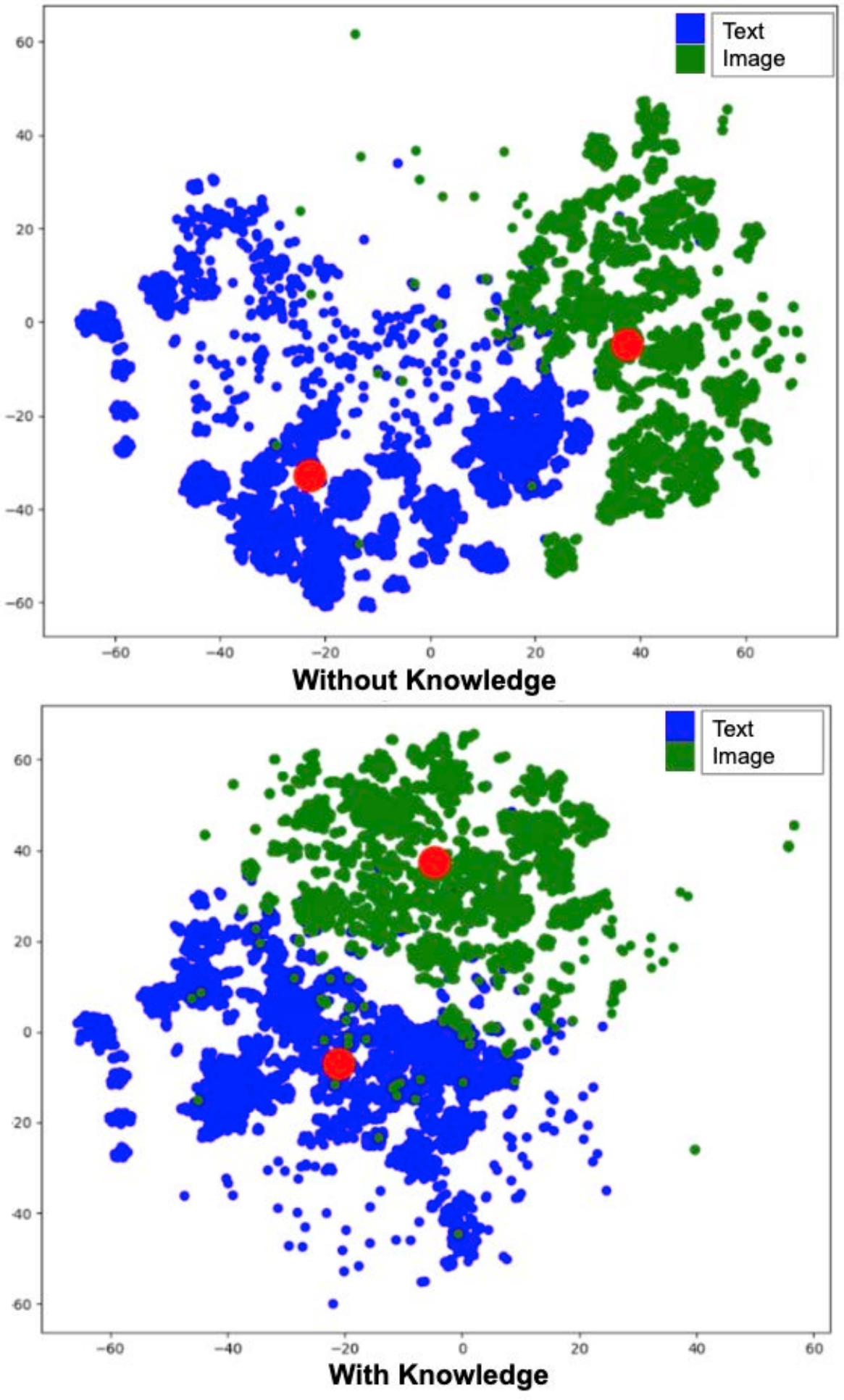}
\vspace{-1.2em}
\caption{t-SNE visualization of text and generated image caption embeddings as two clusters. The red dots are centroids. The two clusters get denser and the distance between them reduces when we include external knowledge.}
\vspace{-1.6em}
\label{fig:embds-cluster}
\end{figure} 

\begin{figure}
  \centering
  \includegraphics[width=8cm]{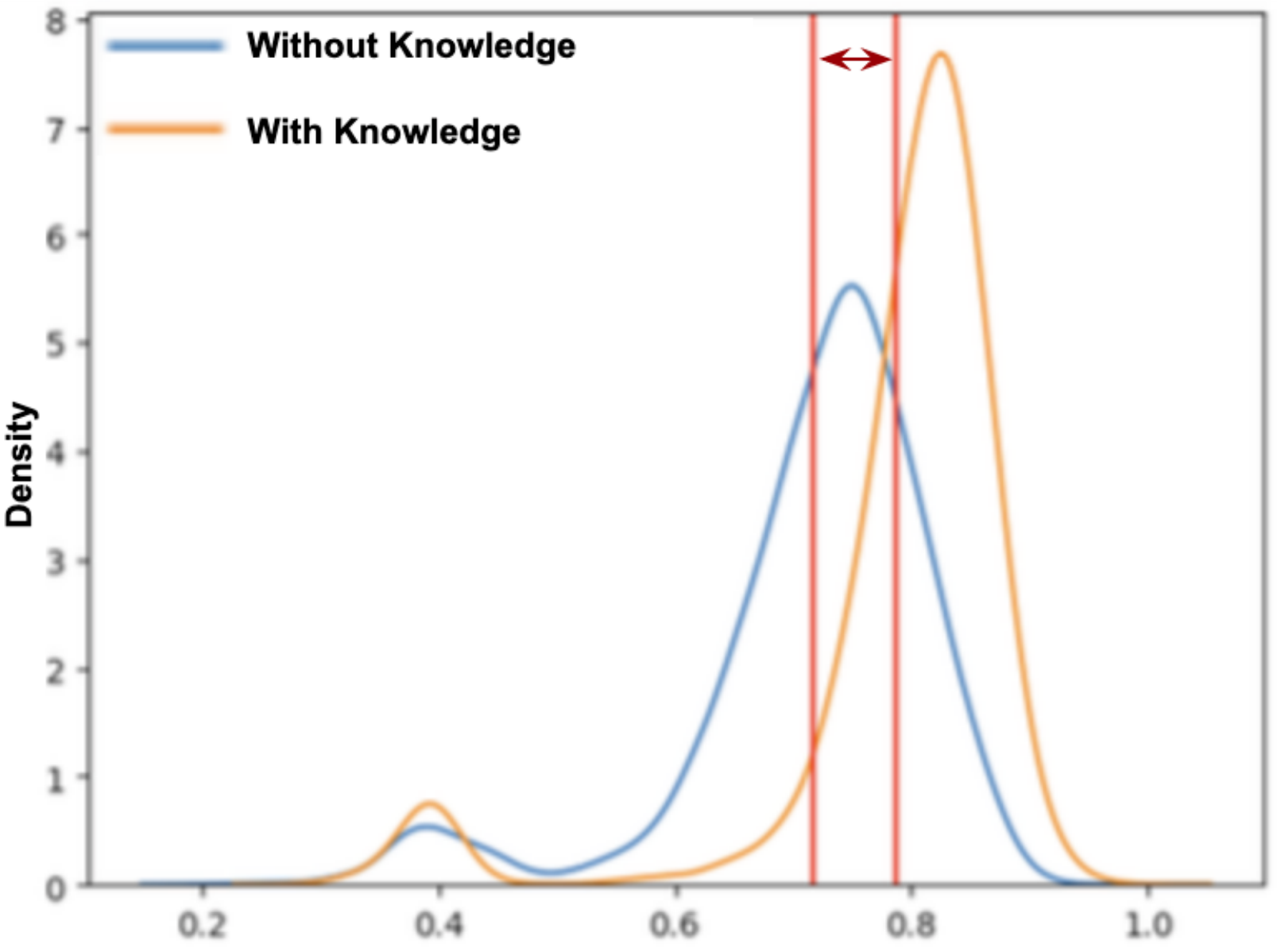}
\vspace{-1em}
\caption{Density Plot demonstrates the difference between the similarities (cosine) of the image and text representations with and without knowledge. The inclusion of knowledge in the input gets these modalities closer by $9.9\%$.}
\vspace{-1.7em}
\label{fig:density-plot}
\end{figure}

Figure \ref{fig:embds-cluster} shows t-SNE visualizations of the text and image (generated captions) clusters created with and without external knowledge. The two clusters with knowledge are denser, and their centroids are closer to each other compared to those without knowledge, demonstrating a reduction in semantic distance. 
We ran the same experiment with different sizes of data points; the clusters consistently got denser and centroids closer (see Figure \ref{fig:context_congruence_across_data_size} in the Appendix).
Further, we measured semantic similarity using cosine for each text-image pair in these clusters, and created density plots, revealing a $9.9\%$ increase with external knowledge in similarity between modalities (see Figure \ref{fig:density-plot}).

In addition, while the contextual relationships between modalities can be recognized to an extent by the VLMs, hallucinations may emerge by generating non-existent connections and non-factual information (see Figure \ref{fig:image-womens-movement-patriachy}). Research shows that external knowledge from KGs can facilitate factual grounding, mitigating hallucinations \cite{agrawal2023can}. We prompted large VLMs, such as LLaVA and BLIP, using pairs of text (actual caption) and image with knowledge. As shown in Table \ref{tab:generated-caption-two-women}, the generated caption with knowledge produces less hallucination compared to those without knowledge. Since incorporating KGs into large foundation models both reduces hallucinations and enhances cross-modal contextual congruence, exploring the association between these improvements presents a promising direction for future research.

These observations demonstrate the potential contribution of KGs and motivate our approach to incorporate them in learning contextually congruent multimodal representations.  

\begin{figure*}[hbt!]
    \centering
  \includegraphics[width=17cm]{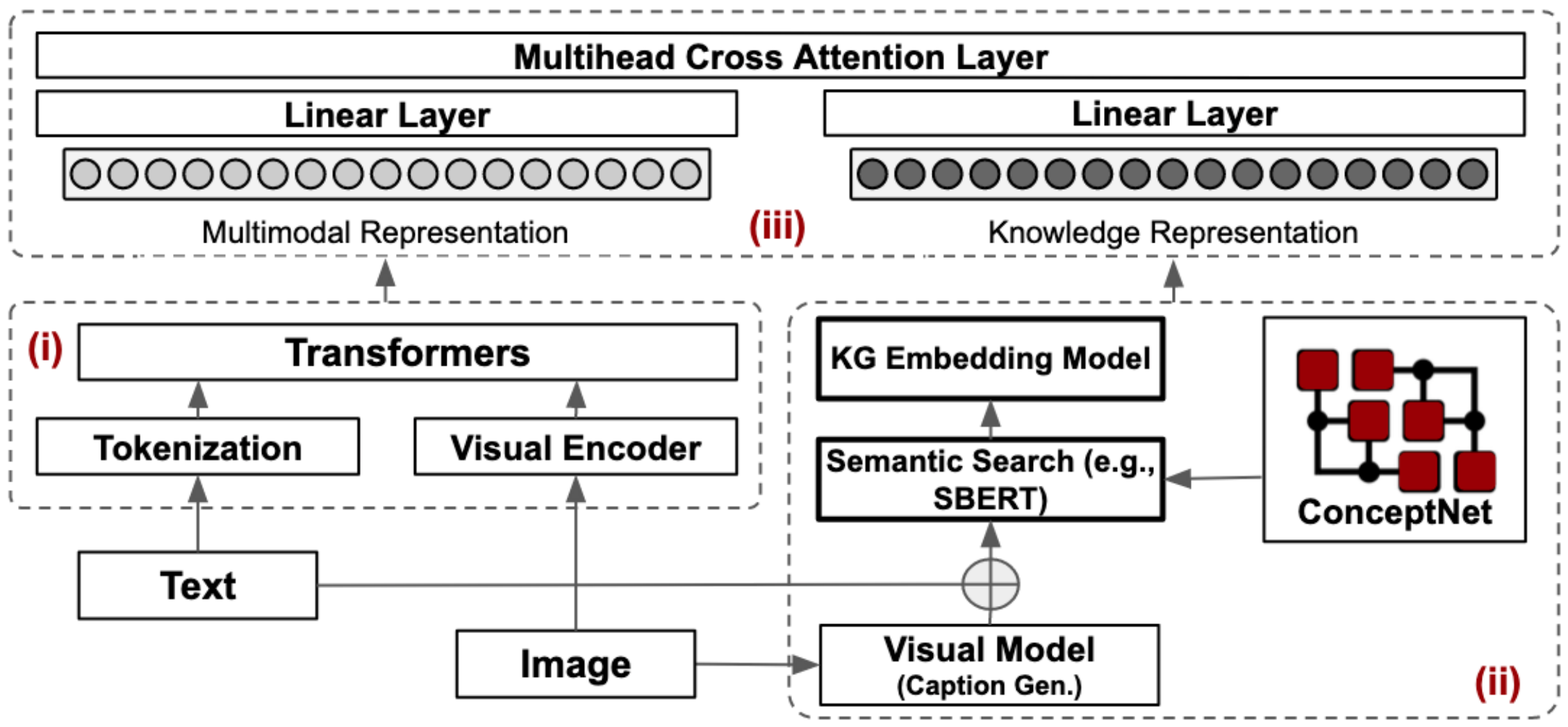}
\vspace{-0.8em}
\caption{Our approach consists of three main components: (i) multimodal learning, (ii) knowledge retrieval and representation, and (iii) knowledge fusion layer. The retrieval component identifies the most relevant concepts from ConceptNet and their KG embeddings are generated. The knowledge fusion layer fuses the multimodal representations with knowledge embeddings followed by Softmax.}
\vspace{-1em}
\label{fig:architecture}  
\end{figure*}

\section{Methodology}
\label{sec:methodology}
\vspace{-1mm}
Our approach comprises of three components (see Figure \ref{fig:architecture}) to predict the success of a crowdfunding campaign, incorporating external knowledge in a multimodal learning framework. We formulate this as a binary classification task, where the outcome is whether a campaign will be successful. First, we used a supervised multimodal learning framework \cite{kiela2020supervised} to learn multimodal representations. Second, we devised a knowledge retrieval component that obtains the most relevant concepts from ConceptNet and generates their knowledge representations using KG Embedding (KGE) models, as described in Section \ref{sec:explicit-knowledge}. Then, we fused the multimodal representation with the knowledge representations to obtain a knowledge-infused representation. This enhanced representation is then input to a classification layer to predict campaign outcomes. 

In this study, we opted to utilize small compact language models, such as BERT and RoBERTa, and visual models, such as ResNet and ViT, which are more accessible in terms of computational cost, yet demonstrated comparable performances. This choice will allow us to improve the performance of compact and lean models as a new competitive baseline for a future study with large VLMs. 

\noindent\textbf{Baselines.} We compared our approach with four different baselines. We used a baseline approach from recent literature for predicting crowdfunding success, MDL-TIM \cite{cheng2019success}. In addition, we used MMBT \cite{kiela2020supervised} as-is, and its two variants with different text and image encoders. 
All models were trained/fine-tuned on the dataset with the same experimental settings (see Section {\ref{implementation-details}).

\subsection{Dataset}
\vspace{-1mm}
We have chosen the Kickstarter crowdfunding platform for its promotional multimodal content, where we have a valid outcome measure with a relatively large N. The dataset was sourced from Webrobots.io \cite{robots2018kickstarter}, covering 75,810 projects seeking funding between June 17, 2021, and May 19, 2022. The dataset will be made available and fully comply with the FAIR principles \cite{force11}. For each project, we tracked indicators such as project ID, name, image, text blurb (used as the text caption in our analysis), project status, goal amount in USD, pledged amount in USD, and industry category. The projects spanned categories including art, comics, crafts, design, film, music, and technology. The primary outcome variable was fundraising success, which was defined in four categories: successful, canceled, failed, and live. Successful projects met their funding goals within the deadline (39\%), while 58\% failed, 5\% were canceled before launch, and 2\% were live at the time of analysis. The canceled and failed projects were grouped as “unsuccessful,” making the outcome variable dichotomous (successful vs. unsuccessful). 
\vspace{-2mm}
\subsection{Problem Formulation}
\vspace{-1mm}
Our dataset contains data points  ($d_i$) consisting of pairs of image ($v_i$) and text ($t_i$) with binary labels ($a_i$); $d_i = {(v_i,t_i,a_i)}$ for $i=1...n$.
Our target classification feature is 
based on whether a campaign is successful; in other words, whether it has reached its funding goal within a given timeline.
Hence, our problem is to determine how likely a given marketing campaign with multimodal content (e.g., text, image) on a crowdfunding platform will be successful.
We denote the knowledge graph as $K_g$ and the set of concepts queried from $K_g$ as $C = {\{c_1, c_2, c_3, ....., c_n\}}$.

\vspace{-2mm}
\subsection{Multimodal Learning}
\label{mmbt-kil-model}
\vspace{-1mm}
Given each pair of image and text ($v_i,t_i$), we followed \cite{kiela2020supervised}, 
because of its modular and flexible framework as the text and image encoders can be replaced and fine-tuned. More specifically, we jointly fine-tuned the pretrained text (e.g., BERT, RoBERTa) and image encoders (e.g., ViT, ResNet) 
using a bidirectional transformer. We initialized the bidirectional transformers with the fine-tuned text encoders and learned weights $W_n \in \mathbf{R}^{P\times D}$,  to map each image ($v_i$) embedding to the token input embedding space, where \( P \) and \( D \) are the dimensionality of the image and the token input embedding space, respectively.

\paragraph{Vision Encoder}
\label{vision-encoder}
We experimented with two types of vision encoders: ResNet-152 \cite{he2015deep} and Vision Transformers \cite{dosovitskiy2021image} model with average pooling over \begin{math} K \times M \end{math} grids in the picture, producing \begin{math} N = KM \end{math} output vectors with a total of 2048 dimensions for each image. As suggested in \cite{kiela2020supervised}, we resized, center-cropped, and normalized the images.

\paragraph{Text Encoder}
\label{text-encoder}

Our approach utilizes base-uncased BERT \cite{devlin2018bert} and RoBERTa \cite{liu2019roberta} as our multimodal framework (MMBT) is based on bi-transformers. Contextual embeddings are created by adding distinct 
$D$-dimensional segments, location, and token embeddings. We learned weights $W_n \in \mathbf{R}^{P\times D}$ to project each of the $N$ image embeddings to $D$ dimensional token input embedding space: 

\begin{equation}
    I_n = W_nf(img,n)
\end{equation}

where $f(·, n)$ is the \textit{n}-th output of the image encoder’s final pooling operation. We used 0-indexed positional coding, i.e., we start counting from 0 for each segment. 

\subsection{Knowledge Embedding Models}
\label{sec:knowledge-embedding-models}

Knowledge graph embeddings, also known as knowledge graph representation learning, are techniques used to represent entities and relationships as dense, low-dimensional vectors. These embeddings capture the semantic relationships and similarities between entities and allow for efficient computation and reasoning in KG-based tasks. In our study we refer to the knowledge graph as $K_g$. Each triple in $K_g$ is represented as $(h,r,t)$ where the entities relation, head and tail are represented as  $h$, $r$ and $t$ respectively.

\noindent \textbf{TransE} \cite{bordes2013translating} is a translational energy-based model that represents entities and relations as vectors of the same Euclidean space $R^d$. For a given triple, score function is defined as $f_r(h,t) = -||h+r-t||_{l_1,l_2}$.

\noindent \textbf{RotateE} \cite{sun2019rotate} is a translational based model that represents entities and relations in complex vector space $C^d$ and defines each relation as an element-wise rotation from the head entity to the tail entity. The score function is $f_r(h,t) = - ||h\odot r - t||$, where $\odot$ is the Hadamard (or element-wise) product. 

\noindent \textbf{DistMult} \cite{yang2014embedding} DistMult is a semantic-matching model, where the score of a triple can be computed as the bilinear product which is represented as $f_r(h,t) = h\times r \times t$ where $e_h,e_t \in R^d$ and $r \in R^{d \times d}$. Here, restricting $R$ to a diagonal matrix, the score is computed as $f_r(h,t) = h\times diag(r) \times t$.


\subsection{Knowledge Retrieval and Representation}
\label{sec:explicit-knowledge}

We generated text captions for images using LVMs to be utilized for knowledge retrieval, along with text (actual caption). BLIP \cite{li2022blip} was observed to perform better than other models and was sufficiently accurate in representing the image content. The text and the generated caption of the image were combined and input into the semantic search module. For semantic search, we utilized Sentence-BERT \cite{reimers2019sentence} to search efficiently and find the most semantically meaningful matches (e.g., $top-k$) between  KG concepts and the input. Sentence-BERT was trained with a regression objective function (e.g., mean squared error) and a siamese network architecture over a sizable natural language inference (NLI) data, which enables efficient inference and semantic search capabilities. Due to its comprehensive coverage of commonsense knowledge, we use ConceptNet \cite{speer2018conceptnet} as our KG. In order to retrieve the most relevant concepts, Sentence-BERT computes the similarity scores, resulting in the $top-k$ concepts ($c_i$) that are semantically similar to the input text and image caption pairs ($v_i,t_i$). In our experiments, we empirically selected $k= 10$. 

As we obtained the most relevant concepts from ConceptNet, we input them into the KGE models that we trained over ConceptNet (see Section \ref{sec:knowledge-embedding-models}) to generate knowledge representations. In our experiments, we used three KGE models: TransE \cite{bordes2013translating}, RotatE \cite{sun2019rotate}, and Distmult \cite{Yang2014-yd}. Note that these KG embeddings capture the overall structure of the graph surrounding the concept of interest, which mainly contributes to the meaning. The generated KG embeddings were 256 dimensions. Then, the knowledge representations were fused with the multimodal representations in the fusion module.


\vspace{1em}
\subsection{Knowledge Fusion}
\label{knowledge-fusion-layer}


We design a knowledge fusion layer using a linear projection to map both multimodal and knowledge representations into a unified common space. Let the multimodal representation be \( M \in \mathbb{R}^{d_m} \) and the knowledge representation be \( K \in \mathbb{R}^{d_k} \). These are projected into a common latent space of dimension \( d_c \) as follows:

\[
M' = W_m M + b_m, \quad K' = W_k K + b_k
\]

\noindent where \( W_m \in \mathbb{R}^{d_c \times d_m} \) and \( W_k \in \mathbb{R}^{d_c \times d_k} \) are learned weight matrices, and \( b_m \in \mathbb{R}^{d_c} \) and \( b_k \in \mathbb{R}^{d_c} \) are the bias terms. To effectively capture the most pertinent aspects of the input from the fusion layer, we employed multi-head cross-attention. The attention mechanism for each head \( i \) is defined as:



\begin{equation}
\begin{split}
h_i &= \text{\small Attention}(QW^Q_i, KW^K_i, VW^V_i) \\
    &= \text{\small softmax}\left(\frac{(QW^Q_i)(KW^K_i)^T}{\sqrt{d_k}}\right) VW^V_i
\end{split}
\end{equation}

\noindent where \( Q \), \( K \), and \( V \) represent the input query, key, and value matrices, and \( W^Q_i \), \( W^K_i \), and \( W^V_i \) are learned projection matrices for the \( i \)-th attention head. The outputs of the attention heads are then concatenated and linearly projected as follows:
\begin{equation}
\text{\small MultiHead}(Q, K, V) = \text{\small Concat}(h_1, h_2, \dots, h_h) W^O
\end{equation}

\noindent where \( h \) is the number of attention heads, and \( W^O \in \mathbb{R}^{h \cdot d_v \times d_c} \) is the output projection matrix. The fused representation \( F \in \mathbb{R}^{d_c} \) is obtained by directly applying layer normalization to the output of the multi-head cross-attention:
\begin{equation}
F = \text{\small LayerNorm}(\text{\small MultiHead}(M', K', M'))
\end{equation}
This allows the model to dynamically integrate the intersecting and complementary aspects of multimodal and knowledge representations, facilitating effective knowledge fusion in the common space. Next, this fused representation \( F \) is passed through a linear layer to project it into the class prediction space:
\begin{equation}
Z = W_F F + b_F
\end{equation}
where \( W_F \in \mathbb{R}^{C \times D} \) is the weight matrix, \( b_F \in \mathbb{R}^C \) is the bias, and \( C \) represents the number of classes. The logits \( Z \) are then passed through a softmax function to produce the predicted probability distribution over the classes:

\begin{equation}
\hat{y} = \text{\small softmax}(Z) = \frac{\exp(Z_i)}{\sum_{j=1}^{C} \exp(Z_j)}
\end{equation}
where \( \hat{y} \) is the predicted probabilities. Finally, we compute the cross-entropy loss \( \mathcal{L} \) between the predicted class distribution \( \hat{y} \) and the true class labels \( y \):

\begin{equation}
\mathcal{L} = - \sum_{i=1}^{C} y_i \log(\hat{y}_i)
\end{equation}

\noindent where \( y_i \) is the true label (encoded as a one-hot vector), and \( \hat{y}_i \) is the predicted probability for the \( i \)-th class. Thus, the model is trained to minimize this cross-entropy loss, to approximate the true label distribution \( y \) in predictions \( \hat{y} \), based on the knowledge-infused multimodal representations.

\subsection{Implementation Details}
\label{implementation-details}
For all models, we swept over the learning rate in ($1e-4$, $5e-5$) with an early stop on validation accuracy for the datasets. The loss function was cross-entropy, and we used BertAdam \cite{devlin2019google} with a warmup rate of $0.1$. The BERT embedding dimension was $768$. The model was trained with a batch size of 16. The dataset was divided into training, validation, and test sets. The training dataset contained $45,810$ samples, and validation and test sets consisted of $15,000$ samples each. We used $Nvidia$ $Quadro$ $RTX$ $6000$ $GPU$.
We trained the KGE models, TransE, Distmult and ComplexE with a learning rate $\lambda$ of  $0.001$ for the stochastic gradient descent, and the latent dimensions $k = 256$. The dissimilarity measure $d$ was set to $L^2$ distance.

\makeatletter
\def\@makecaption#1#2{\vskip\abovecaptionskip \small #1. #2\par \vskip\belowcaptionskip}
\makeatletter

\begin{table*}[hbt!]
  \centerline{%
    \begin{tabular}{p{0.5cm}p{2.1cm}p{2.1cm}p{2cm}p{2cm}p{2cm}p{2cm}p{2cm}}
    \toprule[3pt]
   \# & \textbf{Vision} & \textbf{Language} & \textbf{Knowledge} & \textbf{Precision} & \textbf{Recall} & \textbf{F1} & \textbf{AUC} \\
    \toprule[1pt]
    1 & \multicolumn{2}{c}{MDL-TIM \cite{cheng2019success}} & \centering - & 0.76 & 0.76 & 0.76 & 0.80 \\
    2 & Resnet152 & BERT & \centering -  & 0.86 & 0.77 & 0.81 & 0.86 \\ 
    3 & ViT & BERT & \centering -        & 0.88 & 0.84 & 0.84 & 0.86 \\ 
    4 & ViT & RoBERTa & \centering -     & 0.92 & 0.88 & 0.91 & 0.91 \\ 
    \toprule[1pt]
    5 & ViT & RoBERTa & TransE & 0.92 & \textbf{0.90} \textcolor{green}{(+2.27\%)} & 0.91 & 0.91\\
    6 & ViT & RoBERTa & RotatE & 0.92 & \textbf{0.90} \textcolor{green}{(+2.27\%)} & 0.91 & 0.91\\
    7 & ViT & RoBERTa & DistMult & 0.91 \textcolor{red}{(-1.09\%)} & \textbf{0.89} \textcolor{green}{(+1.14\%)} & 0.91 & 0.90 \textcolor{red}{(-1.10\%)}\\
    8 & ViT & BERT & TransE & 0.92 & 0.85 \textcolor{red}{(-3.41\%)} & 0.89 \textcolor{red}{(-2.20\%)} & 0.90 \textcolor{red}{(-1.10\%)}\\ 
    9 & ViT & BERT & RotatE & 0.92 & \textbf{0.85} \textcolor{red}{(-3.41\%)} & \textbf{0.89} \textcolor{red}{(-2.20\%)} & \textbf{0.90} \textcolor{red}{(-1.10\%)}\\
    10 & ViT & BERT & DistMult & 0.92 & \textbf{0.86} \textcolor{red}{(-2.27\%)} & 0.89 \textcolor{red}{(-2.20\%)} & 0.91\\
    
    11 & Resnet152 & RoBERTa & TransE    & \textbf{0.95} \textcolor{green}{(+3.26\%)}  & \textbf{0.91} \textcolor{green}{(+3.40\%)} & \textbf{0.92} \textcolor{green}{(+1.00\%)} & \textbf{0.94} \textcolor{green}{(+3.00\%)}\\ 
    12 & Resnet152 & RoBERTa & RotatE & \textbf{0.93} \textcolor{green}{(+1.09\%)} & \textbf{0.92} \textcolor{green}{(+4.55\%)} & \textbf{0.92} \textcolor{green}{(+1.10\%)} & \textbf{0.93} \textcolor{green}{(+2.20\%)}\\
    13 & Resnet152 & RoBERTa & DistMult & \textbf{0.95} \textcolor{green}{(+3.26\%)} & \textbf{0.90} \textcolor{green}{(+2.27\%)} & \textbf{0.92} \textcolor{green}{(+1.10\%)} & \textbf{0.93} \textcolor{green}{(+2.20\%)}\\
    14 & Resnet152 & BERT & TransE & \textbf{0.93} \textcolor{green}{(+1.09\%)} & \textbf{0.89} \textcolor{green}{(+1.14\%)} & 0.91 & \textbf{0.93} \textcolor{green}{(+2.20\%)}\\
    15 & Resnet152 & BERT & RotatE & \textbf{0.94} \textcolor{green}{(+2.17\%)} & \textbf{0.89} \textcolor{green}{(+1.14\%)} & \textbf{0.92} \textcolor{green}{(+1.10\%)} & \textbf{0.93} \textcolor{green}{(+2.20\%)}\\
    16 & Resnet152 & BERT & DistMult & \textbf{0.94} \textcolor{green}{(+2.17\%)} & \textbf{0.91} \textcolor{green}{(+3.40\%)} & \textbf{0.92} \textcolor{green}{(+1.10\%)} & \textbf{0.93} \textcolor{green}{(+2.20\%)}\\

    \bottomrule[3pt]
    \end{tabular}%
  }
\caption{Results of the baseline models (\#1-4) and the Knowledge-infused multimodal models (\#5-16), with different text and image encoders and KGE models.}
\vspace{-1.5em}
\label{tab:results}
\end{table*}

\section{Results \& Discussion}
\label{sec:results-discuss}
\vspace{-1mm}
The results of our experiments for the prediction of crowdfunding campaign success is presented in Table \ref{tab:results}. Our models with knowledge-infused representations (\#5-16) provided overall better performance compared to the baseline models (\#1-4). The best performance was achieved with Resnet152 and RoBERTa with the TransE (\#11) KG embeddings with $0.95$, $0.91$, and $0.92$ of precision, recall, and F1-score, respectively. The lowest performance came from a baseline model, MDL-TIM (\#4), while ViT/RoBERTa (\#3) was the most competitive among the baselines. However, knowledge-infused models consistently performed better. In terms of AUC, our best knowledge-infused model (\#11) outperforms other models with $0.94$, indicating a higher true positive rate and a lower false positive rate. The outperformance of our model in AUC also signals potential improvement in the fairness of the model, warranting future work on the fairness assessment of knowledge-infused models. When we compare the models with the same encoders but with or without knowledge, we observe a consistent enhancement in the performance of the models with knowledge. Specifically, the models \#8-10, which utilize ViT and BERT encoders coupled with knowledge representations, outperform the baseline model \#3 with the same encoders but without knowledge. Similar observations can be made between other models, such as models \#14-16 and baseline model \#1. Following up on our research questions, upon these results, we have observed that incorporating external knowledge in learning multimodal representations improves the contextual congruence of the representations of multimodal content, unveiling subtle cross-modal semantic relationships (RQ1). When the knowledge-infused representations were used in our downstream task of predicting the success of crowdfunding campaigns, it provided more consistent and reliable predictive performance (RQ2).

\vspace{-1em}
\begin{figure}[htb!]
  \centering
  \includegraphics[width=8.5cm]{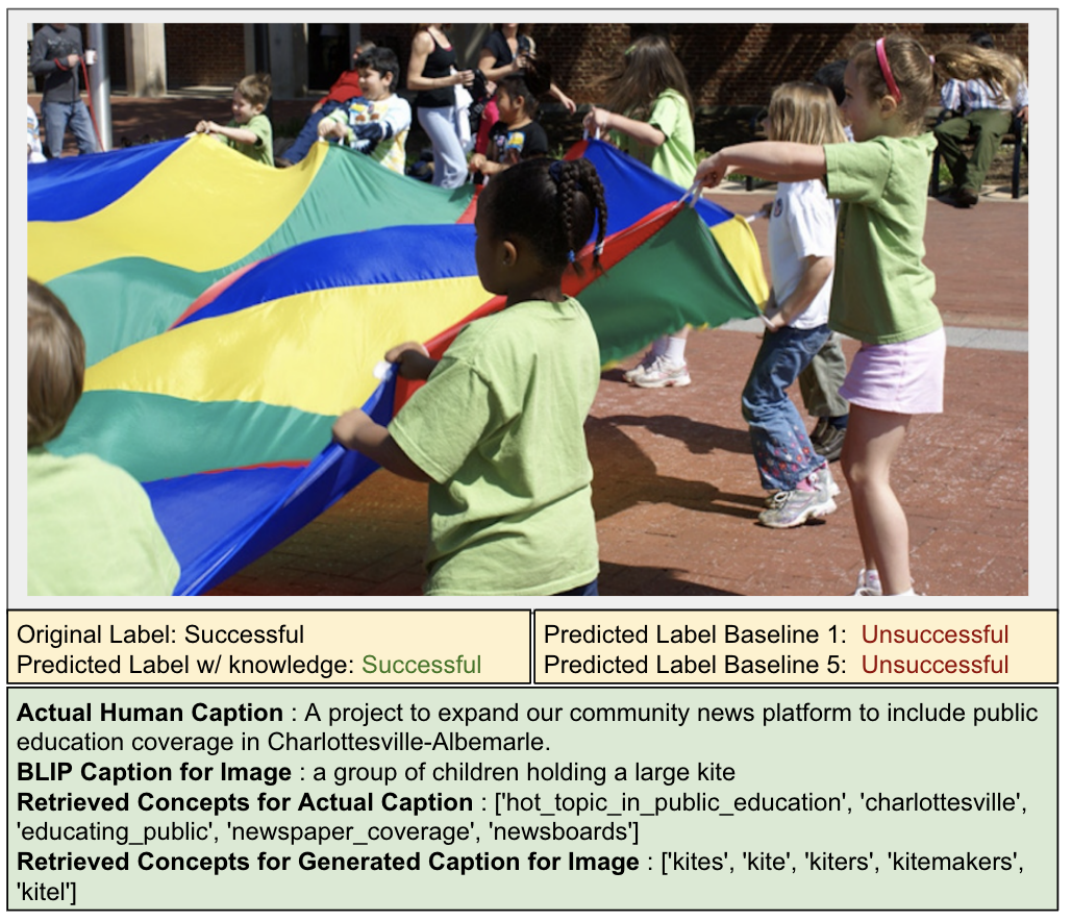}
\caption{The yellow box contains the model prediction with knowledge and two baseline models. The green box shows the actual caption, BLIP caption, and retrieved concepts for each modality.}
\vspace{-1em}
\label{fig:error-analysis-kites-educating-public}
\end{figure} 

\vspace{1mm}
\subsection{Error Analysis}
\vspace{-1mm}
Overall, this analysis aims to explore the contribution of enhanced cross-modal contextual congruence utilizing KGs in multimodal models, as well as limitations compared to baseline models. Hence, we examined a small number of data points where models with knowledge predicted correctly and incorrectly. We make the following observations:

\vspace{1mm}
\noindent \textbf{Erroneous Omissions in Baseline Models.}  In Figure \ref{fig:error-analysis-kites-educating-public}, 
the model with knowledge captures the important concepts or events that connect the image and text to prevent misclassifications and an overall lack of precision. For instance, the baseline models can possibly recognize only objects and shallow details, while additional knowledge and relationships in between provide deeper context into the meaning of the content. For instance, the concept \texttt{\small ‘hot\_topic\_in\_public\_education’} showcases students as \texttt{\small ‘kitemakers’}, \texttt{\small ‘educating\_public’} to support innovative teaching methods by \texttt{\small ‘newspaper\_coverage’}. Such additional richer contextual knowledge helps the model to make better inferences about the meaning of content. Similar limitations can be observed from Figures \ref{fig:error-analysis-poetry-graffiti} and \ref{fig:error-analysis-woofer-error-analysis} in Appendix.

\vspace{1mm}
\noindent \textbf{Erroneous Commissions in the Improved Models.}  The incorporation of noisy knowledge retrieval mechanisms into the model introduces a new set of challenges. In some cases, the model retrieves information that is either irrelevant or contradictory to the context. This phenomenon is particularly prominent when the BLIP model fails to capture the main entities in the context of the image. 
(see Figure \ref{fig:error-analysis-failure-mens-mental-health}) where the campaign talks about how abuse and mental health issues suffered by men are often neglected. The campaign advertisement depicts men in a smaller size and women in a larger size, symbolizing the disproportionate attention given to men’s issues compared to women's. 
As the knowledge retrieved for the image is from ConceptNet through the BLIP caption, many noisy concepts get infused, such as \texttt{\small ‘front\_wall’}, \texttt{\small ‘wall\_placement’}, \texttt{\small ‘front\_walls’}, \texttt{\small ‘recolor\_wall’}. 
This analysis highlights the trade-off between performance enhancement through complex modeling and the susceptibility to noisy knowledge. While the recall improves at the expense of precision with knowledge-infused models, in the case of any misclassifications, the model may predict if a campaign will be successful, which may lead to possible financial loss. Optimization of the knowledge retrieval process to mitigate the incorporation of noisy information can be a possible avenue for future work.

\vspace{1mm}
\noindent \textbf{Social Impact.} Understanding (and thereby predicting) multimodal persuasiveness has extensive application. These are both benign, as in purchasing products or services, or malicious, where prediction is a paramount tool for ensuring safety. In this regard, we note high recall (i.e., low false negatives) without reducing precision (i.e., low false positives), along with higher AUC, warranting further investigation for improvements in fairness using our approach. This distribution of error bears on detection efficiency of critical importance when the cost of error is high and detection requires manual follow-up. While we demonstrate such enhancement for effective multimodal crowdfunding, a similar approach will be necessary for other context-sensitive social and behavioral problems, including toxicity and extremism. Further, the implications of these AI-enabled technologies for individuals and communities, specifically underrepresented groups in society, are still unknown. This work specifically provides potential avenues to address social biases inadvertently encoded in compact and large foundation models (e.g., LLMs, LVLMs) using a framework that incorporates external knowledge in learning for future work.



\begin{figure}
  \centering
  \includegraphics[width=8.5cm]{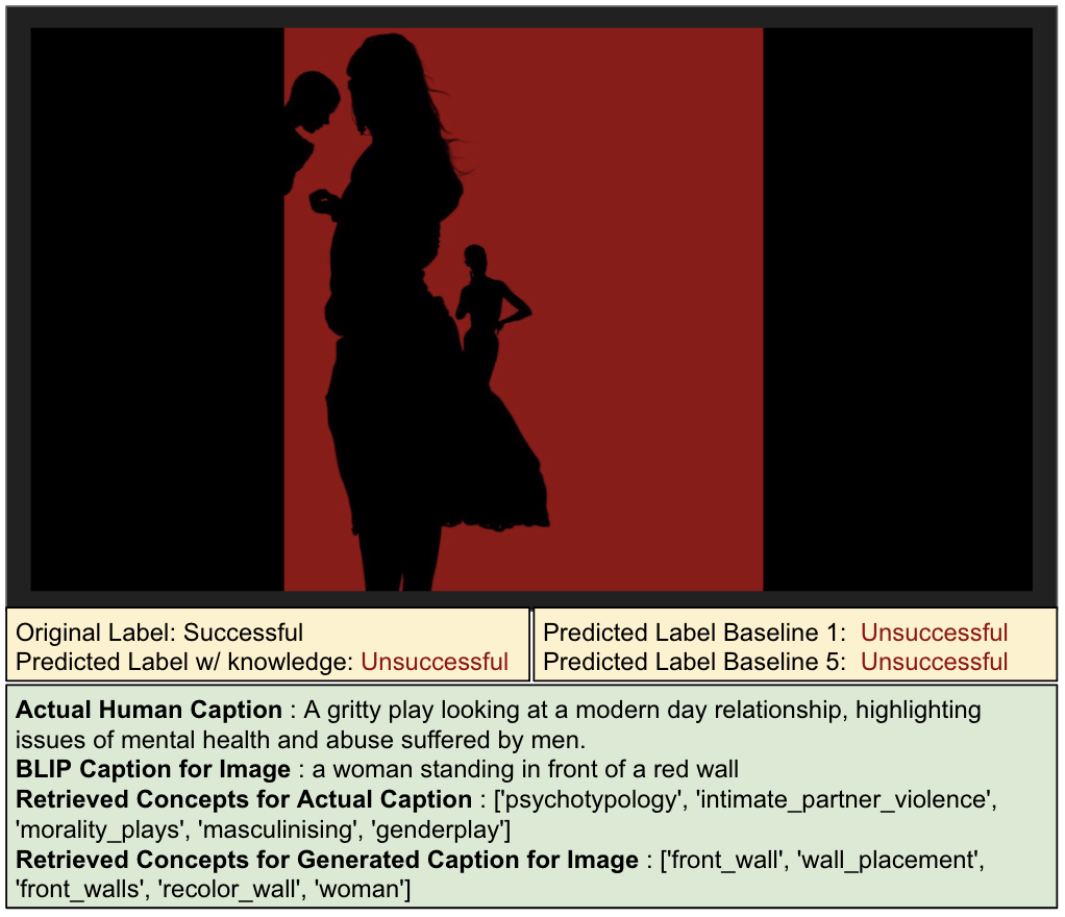}
\caption{In this example, the model with knowledge could not also predict the accurate label along with other baseline models.}
\vspace{-1em}
\label{fig:error-analysis-failure-mens-mental-health}
\end{figure}

\vspace{1mm}
\noindent \textbf{Limitations.} Utilizing KGs to enhance contextual congruence across various modalities shows promise, particularly in crowdfunding marketing. While this approach can be adaptable to various fields, we acknowledge that further research is necessary to validate its effectiveness beyond crowdfunding. We also recognize that the knowledge-infused approach does encounter challenges with noisy triples from the KG, highlighting the need for enhanced robustness. We limited our experiments to the open-source compact models due to the prohibitively high computational costs of large foundation models and left this study for future research. While the approach is applied to crowdfunding, generally a benign domain, it can be misused with malicious intent, such as creating highly persuasive and targeted misinformation campaigns. Therefore, we emphasize the need to follow ethical guidelines and responsible use of these novel approaches to prevent negative outcomes. The data used in this study may contain possible artifacts based on the "country" attribute, which we leave to address in future work. We anonymized our dataset to omit any identifiable information.

\section{Conclusion}
\label{sec:conclusion}
\vspace{-1mm}
Online multimodal crowdfunding campaigns, which involve both images and text, offer an ideal domain for studying the relationship between multiple modalities and predicting campaign success. This setting provides a concrete success metric for evaluating semantic interpretations as effective interpretations accurately predict which image-text pairs will lead to successful campaigns.
In this study, we developed a computational approach to capture the cross-modal context and enhance congruence in multimodal content for crowdfunding campaigns. By incorporating external knowledge from knowledge graphs (e.g., ConceptNet), we enhanced cross-modal contextual congruence in multimodal representations. This incorporation reduced the semantic distance between textual and visual modalities, enabling better capture of cross-modal semantic relationships.
Our findings demonstrate that using external knowledge from knowledge graphs improves the prediction of success in online crowdfunding campaigns over the baseline approaches. This approach not only facilitates the prediction of successful campaigns in the crowdfunding domain but also has broader implications for enhancing the effectiveness of persuasive economic and public service campaigns.

\bibliographystyle{splncs04}
\bibliography{references}


\section{Appendix}
\label{sec:otherappendix}

\begin{table*}[bp]
  \centerline{%
    \begin{tabular}{p{1cm}p{1.5cm}p{13cm}p{1cm}}
    \toprule[3pt]
   \textbf{Model} & \textbf{Knowledge} & \textbf{Generated Caption} & \textbf{Figure \#} \\
    \toprule[1pt]
    BLIP2 & No& a red baseball cap with arabic writing on it & \ref{fig:muslim-cap} \\
    \toprule[1pt]
    BLIP2 & Yes & muslim Americans are a part of our country and that they, too, make America great
    & \ref{fig:muslim-cap} \\ 
    \toprule[1pt]
    LLaVA & No & The image features a red cap with white lettering, likely a "Make America Great Again" hat, positioned on a white surface. The hat has Arabic writing on it, suggesting it is meant to convey a message of inclusivity and support for Muslim Americans. The message on the cap reads, "Muslim Americans also make America great." This serves as a reminder that Muslim Americans contribute significantly to the country and are essential in making America great, just like individuals from various backgrounds. & \ref{fig:muslim-cap} \\ 
    \toprule[1pt]
    LLaVA & Yes & The image features a red cap with white lettering, likely a "Make America Great Again" hat, placed on a white surface. The hat displays Arabic writing, which says, "Muslim Americans also make America great." This message emphasizes the importance of recognizing and appreciating the contributions of Muslim Americans to the nation, reminding people that they, too, play a vital role in making America great. The concept of Americanization refers to the process of adapting to American culture, values, and traditions, and this hat promotes the idea of Muslim Americans actively participating in and enriching the American society. & \ref{fig:muslim-cap} \\ 
    \toprule[3pt]
    BLIP2 & No &  a ruler is sitting on top of a piece of wood
    & \ref{fig:ruler} \\ 
    \toprule[1pt]
    BLIP2 & Yes &  a ruler is sitting on top of a piece of wood
    & \ref{fig:ruler} \\ 
    \toprule[1pt]
    LLaVA & No &  The image features a silver ruler with an internal bit storage compartment, placed against a brown wooden background. The ruler appears to be a TSA-compliant multi-tool, designed for both measuring and carrying small tools securely while in transit. The ruler's unique storage feature allows it to hold bits or other small objects, making it a practical and convenient tool for various purposes.
    & \ref{fig:ruler} \\ 
    \toprule[1pt]
    LLaVA & Yes & The image features a silver ruler with an internal bit storage compartment, designed for carrying and organizing small tools or bits securely. \textbf{This TSA-compliant multi-tool is particularly useful for transporting and organizing military equipment or cargo, as it can easily fit into the compartments of aircraft. The innovative internal storage compartment allows for efficient use of space, making it a very handy tool to carry and transport in unmanned transportation devices or the cargo hold of aircraft}
    & \ref{fig:ruler} \\ 
    \bottomrule[3pt]
    \end{tabular}%
  }
\caption{For the image and text pairs (see Figures \ref{fig:muslim-cap} and \ref{fig:ruler}), the captions were generated by BLIP2 and LLaVA, with and without external knowledge. The knowledge is retrieved from ConceptNet as explained in section \ref{sec:explicit-knowledge} and appended to the prompt input to these models. While the generated captions with knowledge show less hallucinations and more grounding, we leave further investigation of those to the reader as they are not the focus of this study.}
\label{tab:generated-captions}
\end{table*}

\subsection{Example Text \& Image Pairs}
The Figures \ref{fig:muslim-cap} and \ref{fig:ruler}, and Table \ref{tab:generated-captions} provide further example text-image pairs drawn from our dataset, demonstrating similar observations that we described in Section \ref{sec:intro} and Table \ref{tab:generated-caption-two-women}. We generated captions for these example pairs with and without knowledge using large VLMs, such as BLIP2 and LLaVA, provided in Table \ref{tab:generated-captions}.  

\begin{figure}[hbt!]
  \centering
  \includegraphics[width=8.5cm]{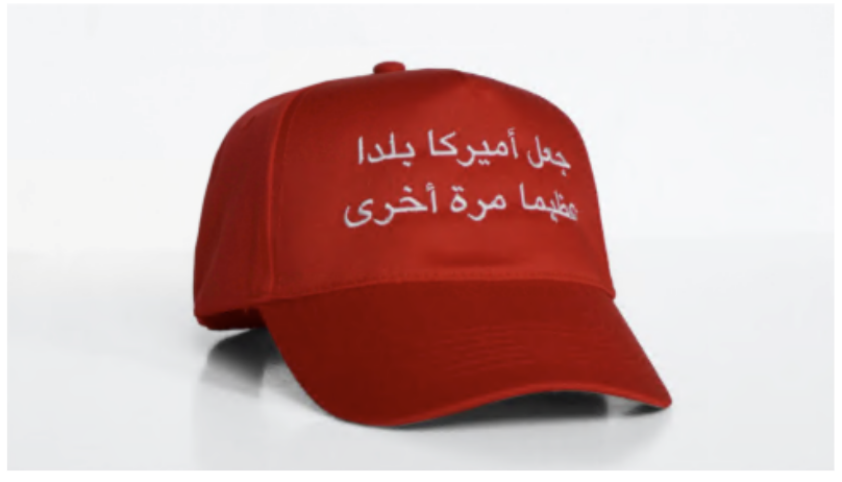}
\vspace{-1em}
\caption{\textbf{Actual caption:} "Let's remind everyone that Muslim Americans are a part of our country and that they, too, make America great."}
\label{fig:muslim-cap}
\end{figure} 

\begin{figure}[hbt!]
  \centering
  \includegraphics[width=8.5cm]{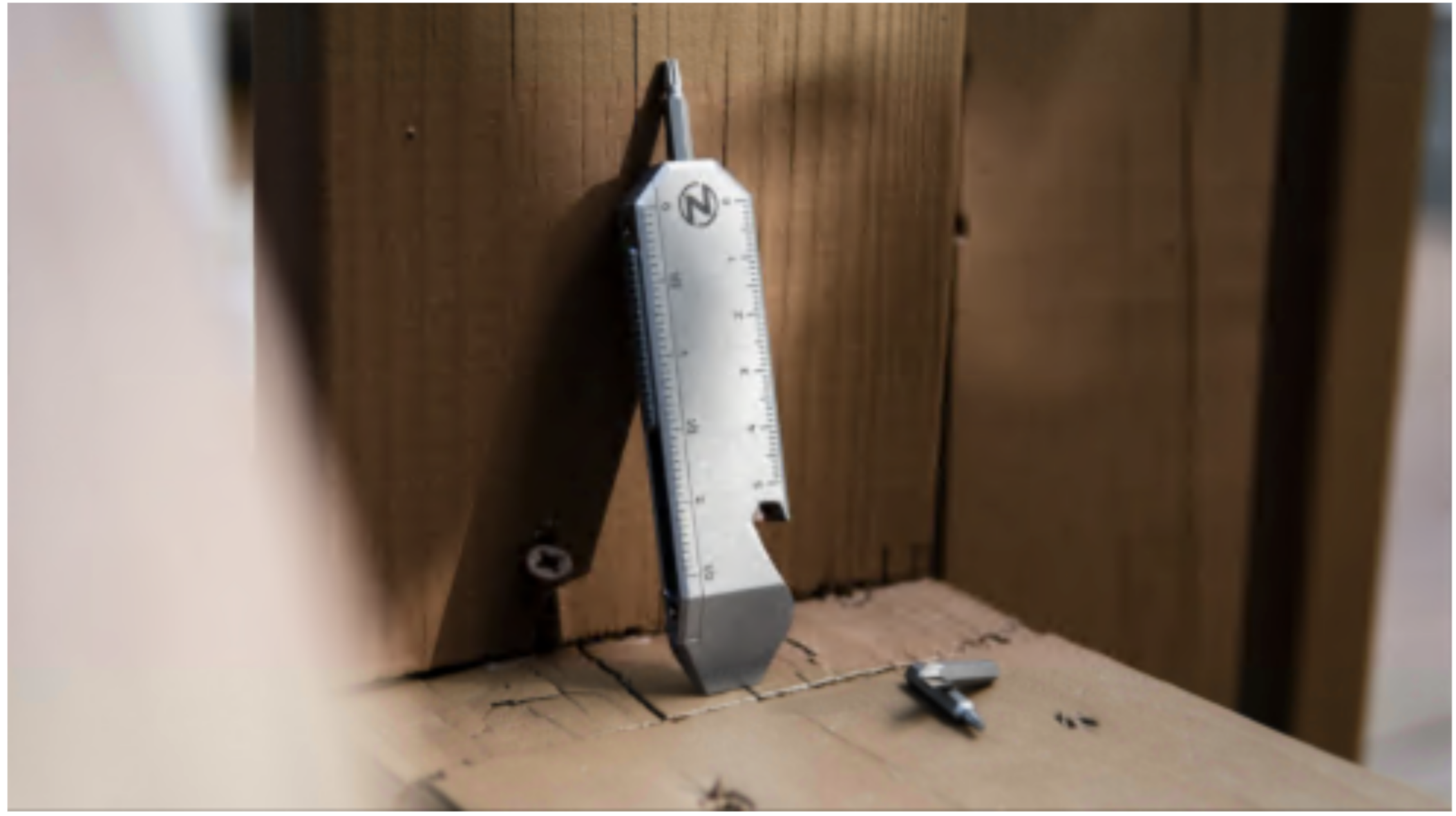}
\caption{ \textbf{Actual caption:} "A TSA-compliant multi-tool with an innovative internal bit storage compartment"}
\vspace{-1.5em}
\label{fig:ruler}
\end{figure} 

\subsection{Extended Exploratory Analysis}
As described in Section \ref{sec:exploratory}, we created clusters for text and image modalities in different sizes of data points to examine if we can make the same observations as shown in Figure \ref{fig:embds-cluster}. When we ran the same experiment with data points ranging from $1,000$ to $20,000$, as shown in Figure \ref{fig:context_congruence_across_data_size}, we observed similar patterns as the clusters got denser and the centroids got closer between them. 

\subsection{Error Analysis Examples}
As we provided two examples for discussing different outcomes from our experiments, this section provides additional examples to contrast the outcomes from knowledge-infused models and models without knowledge for the reader to investigate further. Each example includes figures, actual captions, generated captions, and model outcomes.
For Figure \ref{fig:error-analysis-woofer-error-analysis}, the campaign image promotes a \texttt{ \small ‘subwoofer’} and \texttt{ \small ‘DSP’} solution by depicting a \texttt{\small ‘crowd\_concert’} of fans with raised hands, conveying the message \texttt{\small ‘many\_people\_to\_attend\_concert’.} 
For Figure \ref{fig:music-album-error-analysis}, the campaign ad features a band promoting its music by selling merchandise and offering digital copies on its website. Despite this, noisy entities from ConceptNet, such as \texttt{\small ‘bridge\_distance\_between\_people,’} \texttt{\small ‘bridge, bridges,’ ‘bridgeness,’} and \texttt{\small ‘holding\_up\_bridge’} fail to capture the significant aspects of the image and text pair. For Figure \ref{fig:error-analysis-poetry-graffiti}, the campaign showcases \texttt{ \small ‘performance\_poetry’} combined with \texttt{ \small ‘filmmaking’} highlighting these artistic expressions in unconventional locations like \texttt{ \small ‘graffiti’} on a \texttt{ \small ‘front\_wall’} celebrating both poet performers and \texttt{ \small ‘graffitists’} retrieved from ConceptNet.



\begin{figure}[htbp]
  \centering
  \includegraphics[width=8.5cm, height=8.5cm]{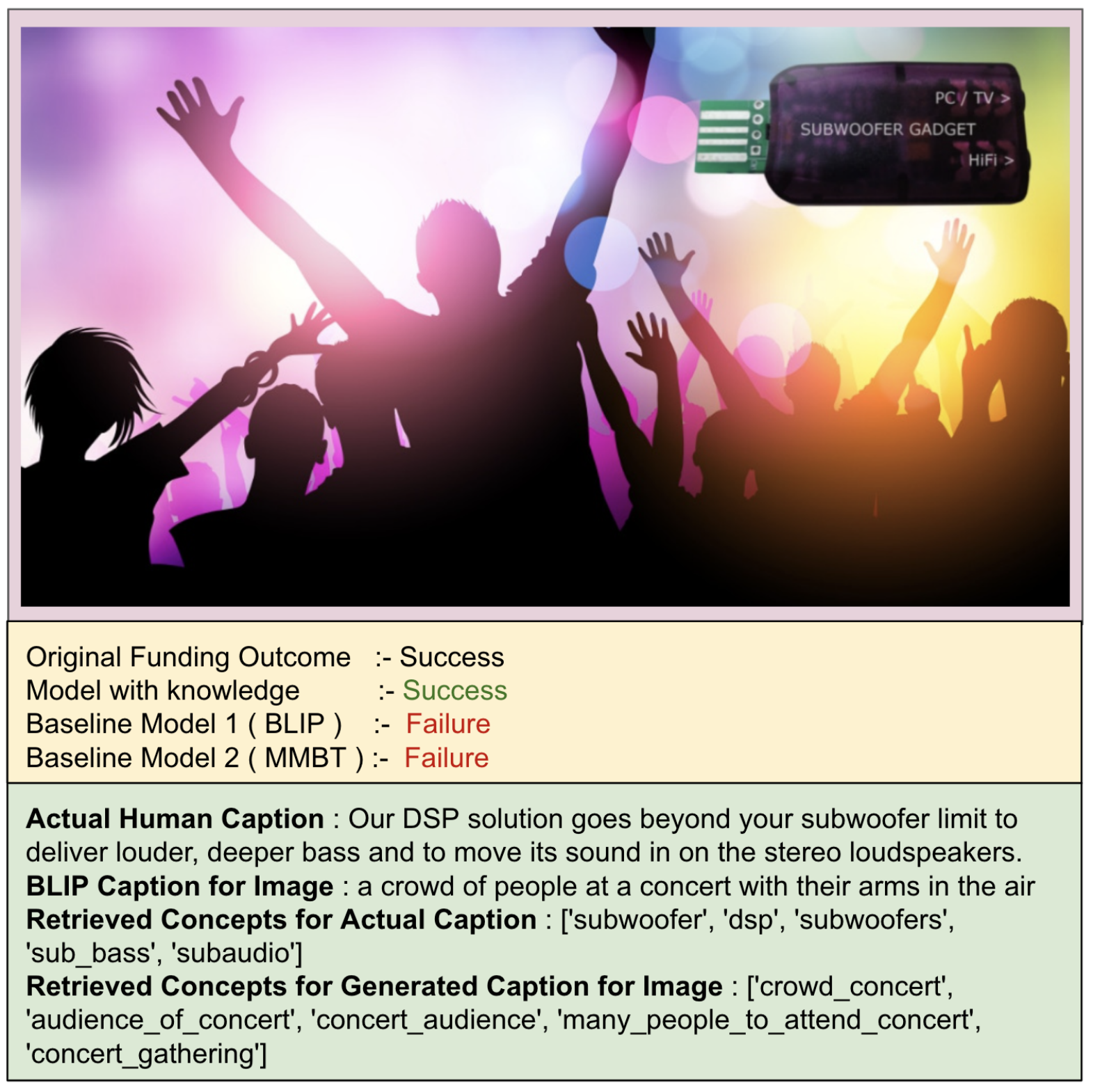}
\vspace{-1em}
\caption{In the yellow box, the prediction of the model with knowledge, BLIP model, and MMBT. In the green box, the actual caption, BLIP caption, and retrieved concepts for each modality.\\
}
\vspace{-1.5em}
\label{fig:error-analysis-woofer-error-analysis}
\end{figure}

\begin{figure}[htbp]
  \centering
  \includegraphics[width=8.5cm, height=8.5cm]{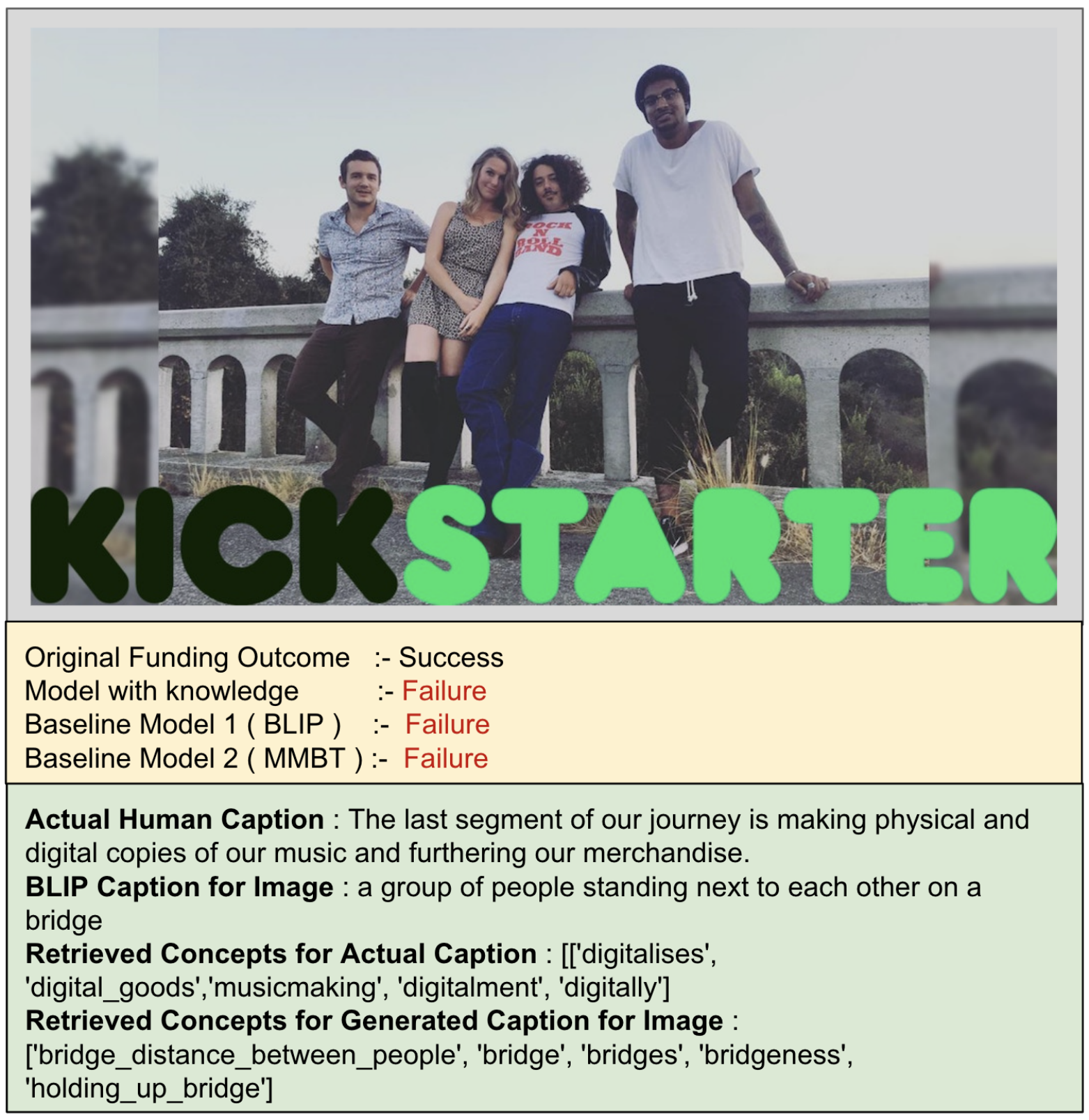}
\vspace{-1em}
\caption{In the yellow box, the prediction of the model with knowledge, BLIP model, and MMBT. In the green box, the actual caption, BLIP caption, and retrieved concepts for each modality.\\
}
\vspace{-1.5em}
\label{fig:music-album-error-analysis}
\end{figure}

\begin{figure}[htbp]
  \centering
  \includegraphics[width=8.5cm, height=8.5cm]{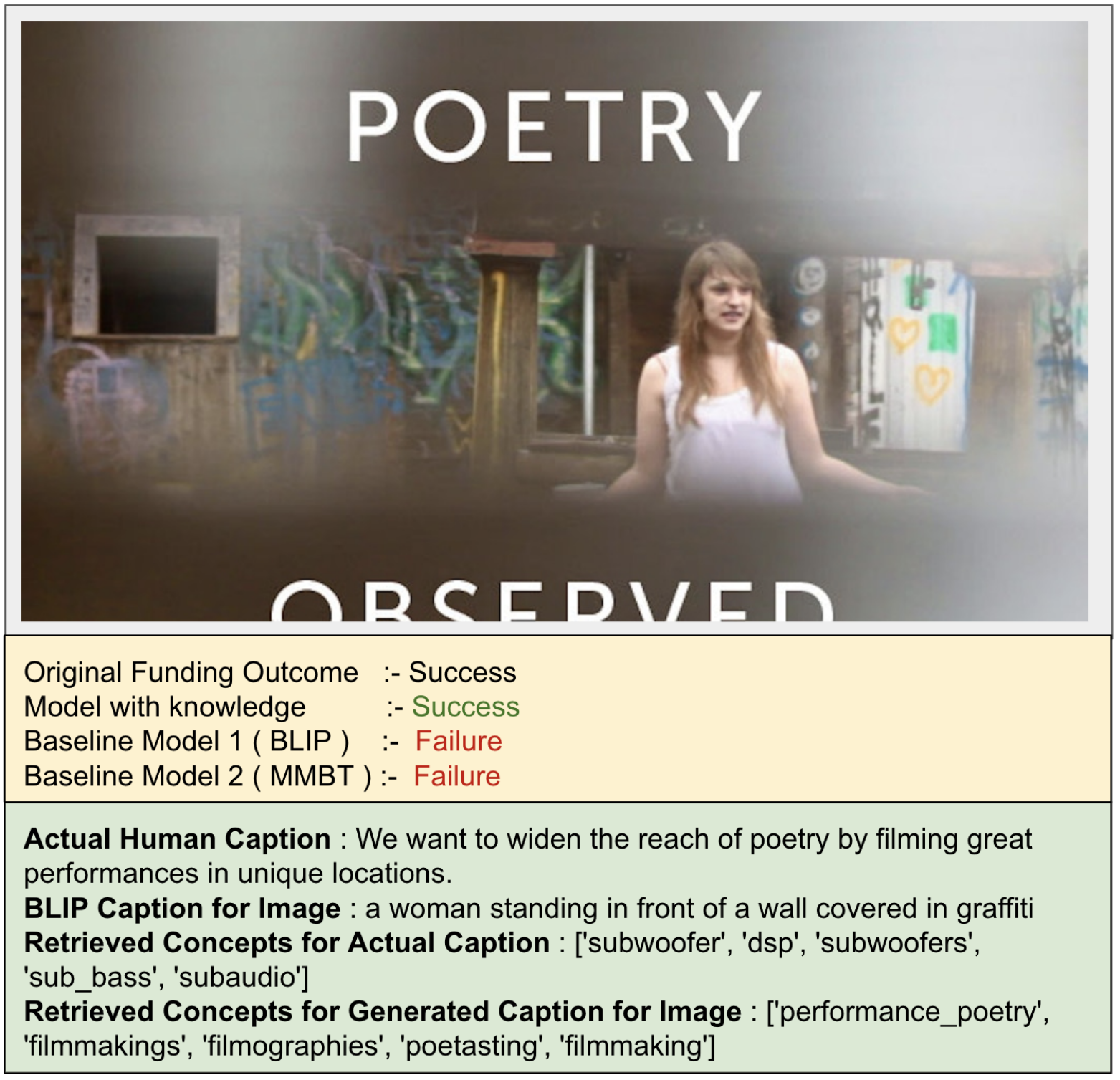}
\vspace{-1em}
\caption{In the yellow box, the prediction of the model with knowledge, BLIP model, and MMBT. In the green box, the actual caption, BLIP caption, retrieved concepts for each modality.\\ 
}
\vspace{-1.5em}
\label{fig:error-analysis-poetry-graffiti}
\end{figure}

\begin{figure}[htbp]
  \centering
  \includegraphics[width=8.5cm, height=8.5cm]{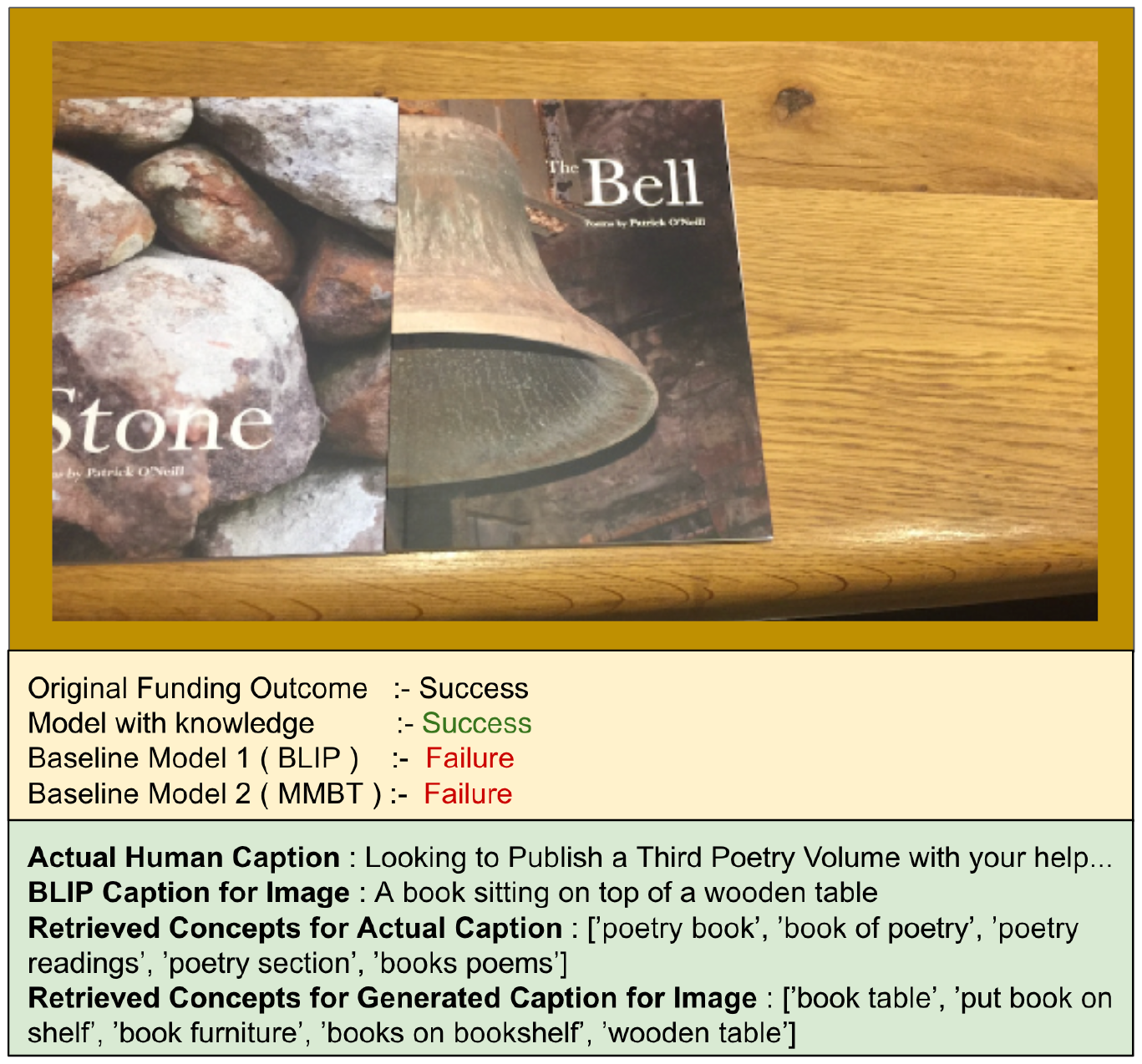}
\vspace{-1em}
\caption{In the yellow box, the prediction of the model with knowledge, BLIP model, and MMBT. In the green box, the actual caption, BLIP caption, retrieved concepts for each modality. \\
}
\vspace{-1.5em}
\label{fig:error-analysis-sucess-poet}
\end{figure} 


\begin{figure*}
  \centering
  \includegraphics[width=18cm]{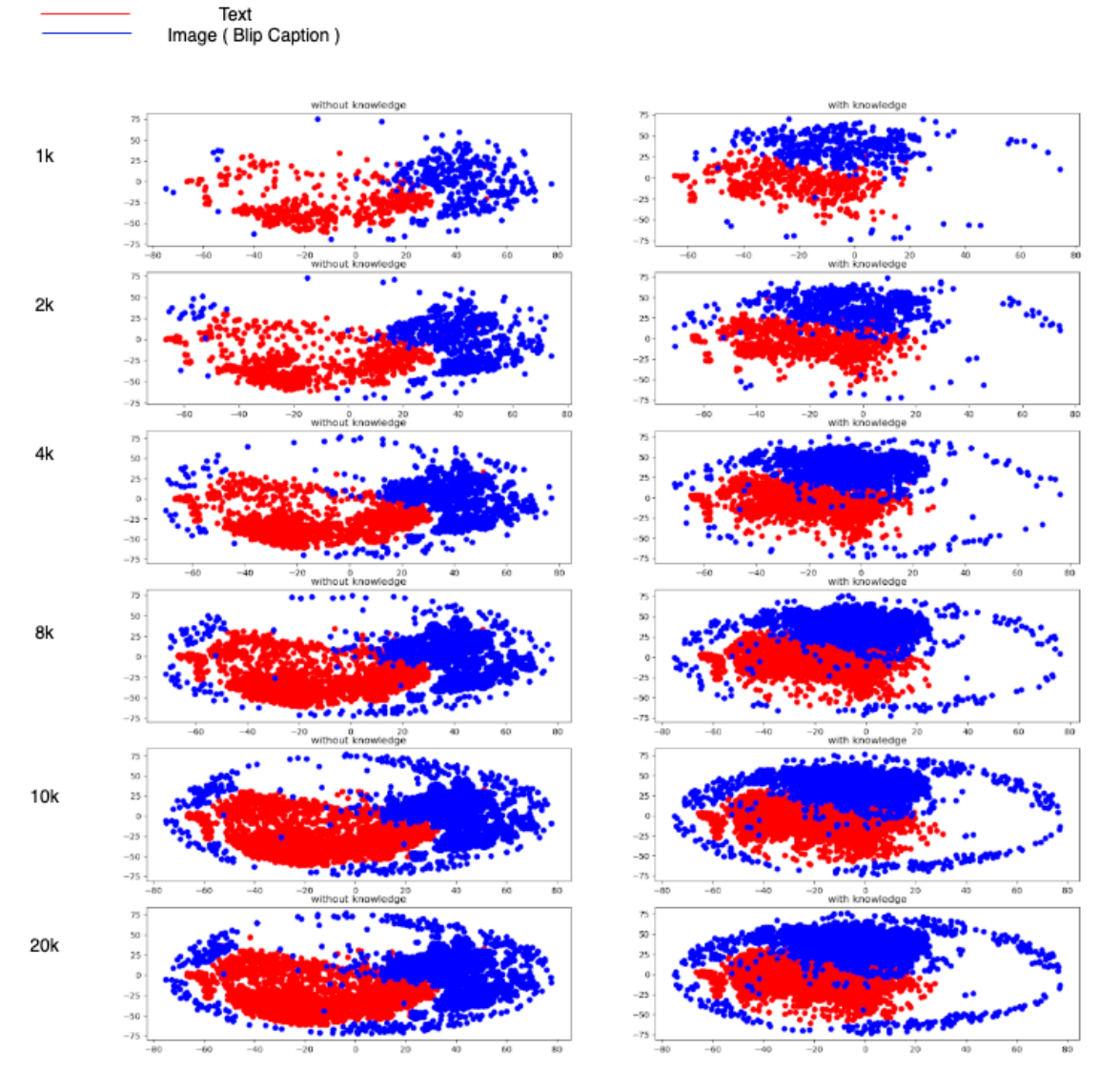}
\caption{Cluster of the embeddings of the BLIP captions and the actual caption embeddings across varying data sizes. The gap between the image and text clusters reduces in the right plot with the addition of knowledge and can be seen across varying data sizes}
\vspace{-1.5em}
\label{fig:context_congruence_across_data_size}
\end{figure*}

\end{document}